**The Afterlives of Shakespeare and Company in Online Social Readership**

Maria Antoniak[1], David Mimno[2], Rosamond Thalken[2], Melanie Walsh[3], Matthew Wilkens[2], Gregory Yauney[2]

[1] Allen Institute for Artificial Intelligence, [2] Cornell University, [3] University of Washington

*All authors contributed equally, and the authors are listed in alphabetical order.*

**INTRODUCTION**

The growth of social reading platforms such as Goodreads and LibraryThing enables us to analyze reading activity at very large scale and in remarkable detail. But twenty-first century systems give us a perspective only on contemporary readers. Meanwhile, the digitization of the lending library records of Shakespeare and Company (SC) provides a window into the reading activity of an earlier, smaller community in interwar Paris. In this article, we explore the extent to which we can make comparisons between the SC and Goodreads communities. By quantifying similarities and differences, we can identify patterns in how works have risen or fallen in popularity across these datasets. We can also measure differences in how works are received by measuring similarities and differences in co-reading patterns. Finally, by examining the complete networks of co-readership, we can observe changes in the overall structures of literary reception.

Both of our data sources consist of interactions between readers and books, but there are substantial differences in context, content, and scale. SC was an English-language bookshop and lending library, run by Sylvia Beach in Paris between 1919 and 1941. Its patrons included famous writers such as Gertrude Stein and Ernest Hemingway, as well as lesser-known writers, artists, and academics. In contrast, Goodreads is a large social website that allows its members to write reviews, to like and comment on others' reviews, rate books, rank books on lists, tag books, participate in reading groups, and take other reading-related actions. Goodreads launched in 2007 and currently has over ninety million members, while the SC dataset contains a total of 6,018 works and 5,601 patrons. Although the SC dataset contains borrowing records for only eleven percent of the library's members, and its records are more comprehensive for the 1930s than for other decades, it nevertheless provides a large amount of information about the reading habits of an important literary community.[1]

    SC and Goodreads differ in other ways, as well. They reflect reading practices separated by nearly a century of historical time. SC catered to a specific community in Paris, while Goodreads users are spread throughout the world. SC readers were mostly bohemian and middle- and upper-class writers, scholars, artists, and critics; Goodreads is used by millions across a presumptively wider (if not necessarily fully representative) socioeconomic range. SC records track the borrowing and purchasing of physical books, while there is no guarantee that a Goodreads reviewer has seen, read, or opened a given book.

    Despite the many factors that distinguish the two datasets, there are also points of commonality that lead us to expect that a comparison would be enlightening. SC was not just any local lending library; it was the focal point of a community of readers that was influential in shaping our contemporary view of the modernist period. Many authors whose works are now considered modernist classics were either members of the community or were personally known by its members. Measuring the specific ways in which SC reading patterns are, and are not, reflected in modern reading patterns may provide insight into the formation of—or at least one version of—the contemporary Anglophone literary canon.

---

[1] Joshua Kotin and Rebecca Sutton Koeser, "The Shakespeare and Company Lending Library Cards in Context," *Shakespeare and Company Project*, Center for Digital Humanities, Princeton University, March 9, 2020.

In this article, we pursue three main strategies for computational comparison of SC and Goodreads. First, we begin by comparing the basic properties of individual works across the two datasets. Of the books that appear in both datasets, we find that, even after a century has passed, almost twenty percent of the hundred most borrowed works in SC are still among the hundred most reviewed books from this set on Goodreads. These enduringly popular books were mostly contemporary to the SC collection; the older books that were favorites of SC readers are less likely to be read and reviewed today. Relatedly, we find that the books that rose the most in popularity from SC to Goodreads include titles by nineteenth-century authors whose work was out of fashion among SC readers (e.g., Jane Austen), while books that fell in popularity include then-contemporary authors (e.g., Dorothy Richardson) who are today primarily the province of scholars. These changes demonstrate the simultaneous expansion and winnowing over time to reach one modern community understanding of "the classics."[2] Our results also suggest that readers today treat many of the modernist-era texts that were first consumed by SC members as an era cohort (that is, as a historically distinct and coherent set) that has remained broadly legible for nearly a century.

Second, we consider the two communities from a network perspective, comparing works on the basis of their connection to other works. We find that many of the books that retain popularity from SC to Goodreads also maintain the same patterns of co-readership between the two groups. This fact again suggests that, at least in the case of books read by SC and Goodreads members, the literary contexts of reception can be stable across widely diverse environments.

---

[2] Melanie Walsh and Maria Antoniak, "The Goodreads 'Classics': A Computational Study of Readers, Amazon, and Crowdsourced Amateur Criticism," *Journal of Cultural Analytics* 6, no. 2 (2021): 243–287; John Guillory, *Cultural Capital: The Problem of Literary Canon Formation* (Chicago, IL: University of Chicago Press, 1993).

Third, we consider the structure of the two networks as a whole. We find that, rather than showing which books were most central, or core, to each community, network analysis magnifies the reading habits of two prolific SC readers (and friends) in a way that may be useful for newly directed historical research.

**DATA COLLECTION AND PROCESSING**

We match 4,460 (74.1 percent) of the 6,018 SC books to records in Goodreads. We use the Goodreads API to search for the book IDs corresponding to each title and author. We then manually check each match and update the book IDs when no match is found or when the matched book is incorrect. Many books have multiple versions in Goodreads, and we prioritize the most specific (e.g., individual volumes rather than collected works) and popular (by numbers of ratings and reviews). From the ranked set of results for each book, we select the book with the highest number of ratings as our final match.

After matching each SC book with its Goodreads book ID, we scrape the Goodreads page for that book. This data includes the ISBN, the number of pages, number of reviews, number of ratings, average and distribution of the ratings, original publication year, and the shelves, genres, and lists to which the book is assigned by Goodreads members. After scraping, we have a set of 4,454 books by 1,726 authors, with 3,940 of those books receiving at least one rating and 3,223 books receiving at least one review. We release this connected metadata at https://github.com/gyauney/shakespeare-and-company-social-readership.

**COMPARING POPULARITY IN SC AND GOODREADS**

To measure the relative popularity of books in the two datasets, we count the number of times the book was borrowed from SC and the number of text reviews the book received on Goodreads. As shown in the plots below, these popularity metrics have a much wider range for Goodreads (maximum reviews: 77,817) than for SC (maximum borrows: fifty-six). Since many SC books have the same (small) number of borrows, fine-grained comparisons of popularity rank are possible only in the Goodreads data. In all the following results, we show only books that appeared in both datasets (according to the matching process described above) (fig. 1).

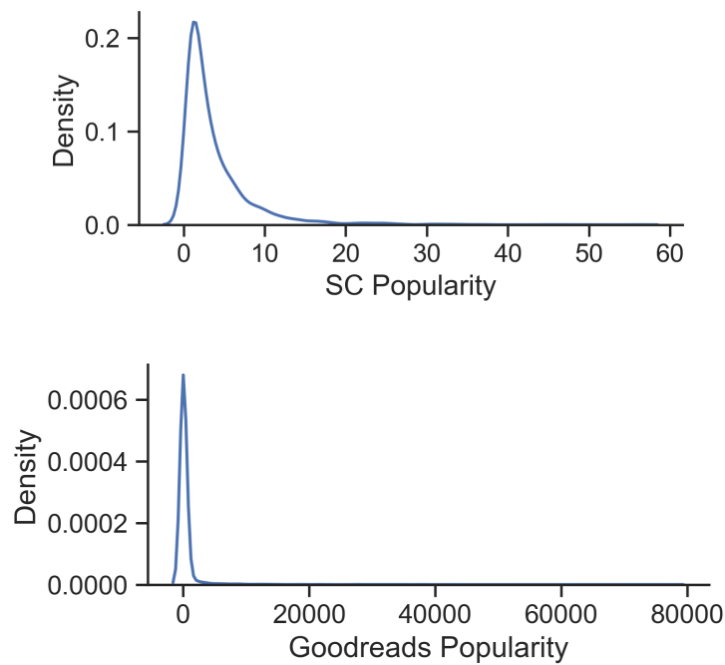

**Fig. 1.** Kernel density estimate (KDE) plots of the popularity distributions for SC and Goodreads.

We begin by listing the most popular books in each of the datasets. James Joyce, Dorothy Richardson, and Katherine Mansfield dominate the SC most popular books—all authors contemporary to SC—while Jane Austen, Charlotte Brontë, and F. Scott Fitzgerald (an SC contemporary and, briefly, member, but not a particular community insider), dominate the Goodreads most popular books. The lists of most popular authors in the two datasets reflect the same pattern. One difference is that D. H. Lawrence is the most borrowed author in SC, despite having only one top-ten most-borrowed book.

| Title | Author | Borrowed |
|---|---|---|
| *A Portrait of the Artist as a Young Man* | James Joyce | 56 |
| *Dubliners* | James Joyce | 47 |
| *Pointed Roofs* | Dorothy M. Richardson | 45 |
| *The Garden Party and Other Stories* | Katherine Mansfield | 37 |
| *Mr. Norris Changes Trains* | Christopher Isherwood | 36 |
| *A Passage to India* | E. M. Forster | 35 |
| *Mrs. Dalloway* | Virginia Woolf | 34 |
| *Women in Love* | D. H. Lawrence | 33 |
| *Point Counter Point* | Aldous Huxley | 33 |
| *The Good Earth* | Pearl S. Buck | 33 |

**Table 1.** The books borrowed most often in the SC dataset. (Only books also included in the Goodreads dataset are shown here).

| Title | Author | Text Reviews |
|---|---|---|
| *Pride and Prejudice* | Jane Austen | 77,817 |
| *The Great Gatsby* | F. Scott Fitzgerald | 75,296 |
| *Jane Eyre* | Charlotte Brontë | 47,139 |
| *Wuthering Heights* | Emily Brontë | 42,403 |
| *And Then There Were None* | Agatha Christie | 40,393 |
| *Of Mice and Men* | John Steinbeck | 37,837 |
| *The Picture of Dorian Gray* | Oscar Wilde | 37,248 |
| *Little Women* | Louisa May Alcott | 34,878 |
| *Brave New World* | Aldous Huxley | 34,579 |
| *Dracula* | Bram Stoker | 29,583 |

**Table 2.** The books reviewed most often in the Goodreads dataset. (Only books also included in the SC dataset are shown here).

| Top SC Authors | Total Borrows (Across Titles) | Top Goodreads Authors | Total Text Reviews (Across Titles) |
|---|---|---|---|
| D. H. Lawrence | 300 | Jane Austen | 168,543 |
| Virginia Woolf | 249 | Agatha Christie | 125,986 |
| Aldous Huxley | 235 | F. Scott Fitzgerald | 84,789 |
| Dorothy M. Richardson | 198 | Charles Dickens | 69,609 |
| W. Somerset Maugham | 187 | John Steinbeck | 60,344 |
| James Joyce | 182 | William Shakespeare | 59,935 |
| Ernest Hemingway | 182 | Oscar Wilde | 53,992 |
| John Galsworthy | 180 | Charlotte Brontë | 53,796 |
| Henry James | 169 | Fyodor Dostoyevsky | 49,215 |
| William Faulkner | 163 | Emily Brontë | 42,503 |

**Table 3.** The most popular authors in the SC and Goodreads datasets.

In order to reduce bias toward the most popular works, we next convert our popularity metrics to ranks, where the 0-ranked book or author is the most popular. We scale these ranks to a [0, 1] range and compare the ranks in SC and Goodreads in the following plots. According to these ranks, the popularity of both the books and authors is correlated across the SC and Goodreads datasets; if a book or author is popular on Goodreads, it is also likely to be popular in SC. But there are many outliers, indicating cases where a book or author is much more popular in one dataset than in the other. These outliers represent a rise or fall in popularity rank. Of the shared books, eighteen of the hundred most borrowed SC books are among the hundred most reviewed Goodreads books (of the books included in our datasets).

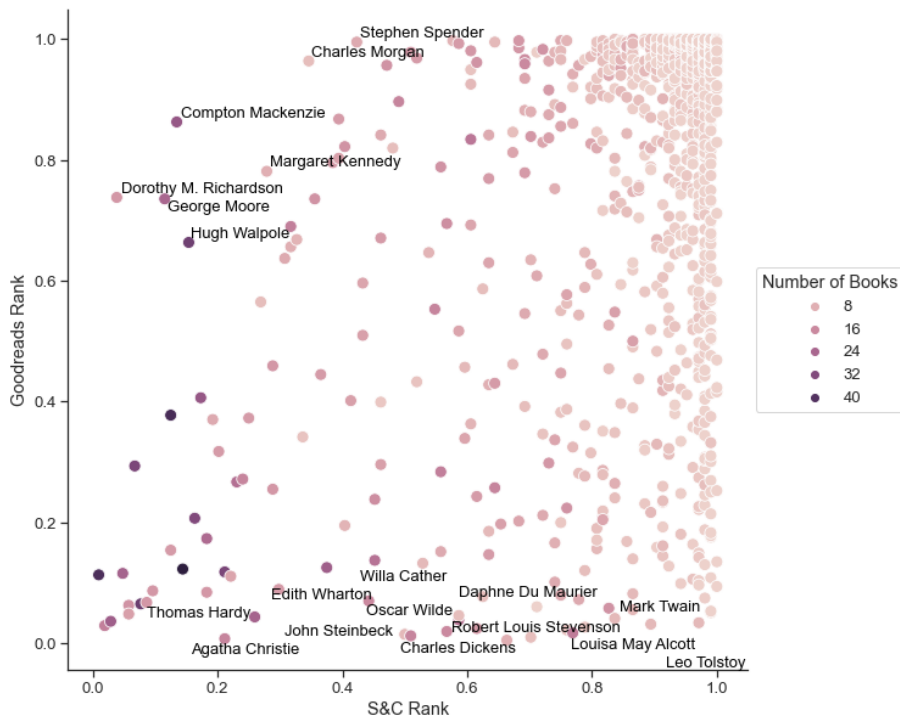

**Fig. 2.** A comparison of author popularity in Goodreads and SC. Each point represents an author, while the x-axis represents the ranked popularity in SC (by number of borrows) and the y-axis represents the ranked popularity in Goodreads (by number of text reviews). We sum the popularity metrics across books for each author. Popularity in Goodreads and SC is correlated (Pearson r=0.51, p<0.05), but there are many outliers representing authors that are much more popular in one dataset than the other. Leo Tolstoy and Louisa May Alcott, for example, are much more popular in Goodreads than they were among SC patrons.

**ROSE THE MOST IN RANK (FROM SC TO GOODREADS)**

To tease apart which books were popular in Goodreads in comparison with SC, we calculate each book's change in rank: its Goodreads rank subtracted from its SC rank. Of the books that rose the most in popularity on Goodreads, many are older books—published before SC opened.

| Change in Rank | Title | Author |
| --- | --- | --- |
| +0.96 | *Little Women* | Louisa May Alcott |
| +0.96 | *Dracula* | Bram Stoker |
| +0.96 | *The Wonderful Wizard of Oz* | L. Frank Baum |
| +0.95 | *Anna Karenina* | Leo Tolstoy |
| +0.92 | *Macbeth* | William Shakespeare |
| +0.92 | *The Trial* | Franz Kafka |
| +0.92 | *Madame Bovary* | Gustave Flaubert |
| +0.91 | *All Quiet on the Western Front* | Erich Maria Remarque |
| +0.91 | *Uncle Tom's Cabin* | Harriet Beecher Stowe |
| +0.91 | *Mansfield Park* | Jane Austen |

**Table 4.** The books that rose the most in popularity rank from SC to Goodreads.

**FELL THE MOST IN RANK (FROM SC TO GOODREADS)**

Of the books that fell most in rank, all were contemporary to the active years of SC. While many of these authors are still well known in academic or literary contexts (e.g., Katherine Mansfield), they are not popular with the majority of readers on Goodreads. For example, even in the 1930s, the modernist and experimental author Dorothy M. Richardson did not enjoy mainstream popularity, but her books were borrowed fairly often by SC patrons, who would have been reading her work alongside contemporaries such as James Joyce and Virginia Woolf. Richardson was also in personal contact with Sylvia Beach. But her work is today largely ignored within the Goodreads community.

| Change in Rank | Title | Author |
| --- | --- | --- |
| -0.59 | *Bliss & Other Stories* | Katherine Mansfield |
| -0.60 | *The Tunnel: Pilgrimage, Volume 4* | Dorothy M. Richardson |
| -0.63 | *Pointed Roofs, Backwater, Honeycomb* | Dorothy M. Richardson |
| -0.63 | *Pilgrimage: Backwater* | Dorothy M. Richardson |
| -0.63 | *The Fountain* | Charles Morgan |
| -0.63 | *Experiment in Autobiography: Discoveries & Conclusions of a Very Ordinary Brain (Since 1866)* | H. G. Wells |
| -0.64 | *Studies in the Psychology of Sex* | Havelock Ellis |
| -0.70 | *South Wind* | Norman Douglas |
| -0.74 | *Sparkenbroke* | Charles Morgan |
| -0.85 | *Pointed Roofs* | Dorothy M. Richardson |

**Table 5.** The books that fell the most in popularity rank from SC to Goodreads.

**LISTS: CHANGE IN RANK (SC TO GOODREADS)**

The previous section shows that there are substantial differences in interactions for both works and authors between the two datasets. But Goodreads also includes features that allow us to explore how readers perceive these changes; these lists can indicate why certain titles rose or fell in popularity. Goodreads lists are ranked sets of books, created by users and to which any user may add any books. Users influence rankings on these lists by voting individual books up or down. These lists are frequent sources of both creative experimentation and the codification of community values.

Examining the lists that Goodreads users assign to the SC books, we find strong evidence of the canonical status of many SC titles. The lists to which SC books are most frequently assigned by Goodreads users are lists of best books (e.g., Best Books Ever, Best Books of the 20th Century) and lists of books everyone should read (e.g., Books That Everyone Should Read At Least Once, Must Read Classics).

| List Name | Number of Books on List |
| --- | --- |
| Best Books Ever | 465 |
| Books That Everyone Should Read At Least Once | 312 |
| 100 Books to Read in a Lifetime: Readers' Picks | 239 |
| Best Books of the 20th Century | 217 |
| 1001 Books You Must Read Before You Die | 212 |
| Must Read Classics | 188 |
| The Great Classics You Have Not Read Yet | 170 |
| The Guardian's "1000 Novels Everyone Must Read" | 168 |
| Best Books of the 19th Century | 164 |
| Books that Blew Me Away and that I Still Think About (of all types) | 162 |

**Table 6.** The most popular lists to which Goodreads users assign books matched with SC.

However, popular books are more likely to be added to lists, and so unpopular books are missing from these categorizations. What about the books that were popular among SC patrons but are not as popular on Goodreads? Did SC members only read what they themselves would have recognized as the "best books" and "books everyone should read," or was their reading more self-consciously avant-garde in historical context?

To answer these questions, we rank the lists by their change in mean rank from SC to Goodreads. For each list, we represent the rank as the mean of the ranks of the books that have been assigned to that list by at least five users. We also require that at least ten books in our matched dataset be assigned to each list; other lists are discarded. The results show which lists are associated with books that rose or fell in popularity from SC to Goodreads. For example, books assigned to lists associated with particular modernist authors (e.g., James Joyce Reading List, Best of D. H. Lawrence) fell in popularity, while books associated with what are today considered "classics" and with children (e.g., Proliferation of the Classics, My Favorite Childhood books) rose in popularity. Evidence from these lists indicates that books that rose in popularity became part of the canon, while books that fell in popularity were particular to the time and place of SC.

| Change in Rank | Lists That Rose Most in Rank | Change in Rank | Lists That Fell Most in Rank |
|---|---|---|---|
| +0.84 | 100 Must Read Books | -0.05 | Interwar British Vogue Recommends... |
| +0.84 | Clean | -0.05 | Best of Sinclair Lewis |
| +0.83 | Books that reached 1000 editions (or more) | -0.05 | James Joyce Reading List |
| +0.83 | Half a million ratings to a million ratings | -0.06 | REALLY Seriously Underrated Books (100 to 500 Ratings) |
| +0.82 | Amazing books that won't make you blush, squirm, get sick or have nightmares! | -0.07 | Best of George Bernard Shaw |
| +0.81 | Proliferation of Classics | -0.07 | Modernism - An Alternate Canon |
| +0.81 | Books Every Child Should Read | -0.08 | Books banned in Ireland 1928-1929 |
| +0.81 | ONE DAY Best Summer Reads | -0.08 | REALLY Underrated Books (Fewer than 1,000 Ratings) |
| +0.80 | Best Free eBooks | -0.08 | Best of D. H. Lawrence |
| +0.80 | My Favorite Childhood books | -0.12 | Underrated Bestsellers, Fewer Than 100 Ratings |

**Table 7.** Lists that rose and fell the most in rank from SC to Goodreads.

**SHELVES: CHANGE IN RANK (FROM SC TO GOODREADS)**

Similar to lists are Goodreads shelves. Where lists are ranked, shelves function more like free-text tags. Users employ shelves both for personal tracking (e.g., to-read, my-favorites, read-in-2020) and to help build the community's mapping of books. In particular, shelves, unlike lists, often function as genre labels (e.g., romance, historical fiction) on Goodreads, and so their examination can reveal new perceptions and aspects of the books that rose or fell in popularity.[3]

As we did with lists, we can compare the shelves associated with books that rose and fell the most in popularity. Again, we measure the rank for each shelf as the mean of the ranks of the books assigned to that shelf by at least five users and discard shelves that were assigned to fewer than ten books.

We find that shelves associated with books that rose the most in popularity generally mention genres (thriller-mystery, chick-lit, childrens-lit), classics (the-classics), or school (read-in-school), themes that prior work has connected to popular discussions of canonization (Walsh and Antoniak, "The Goodreads 'Classics'"). Shelves associated with books that fell in popularity tend to mention specific authors (joyce, d-h-lawrence) or literary criticism (lit-crit, literary-criticism). These results indicate that books that were read in school or that fit genre classifications have remained popular, while books that do not fit genre classifications have decreased in popularity.

---

[3] Maria Antoniak, Melanie Walsh, and David Mimno, "Tags, Borders, and Catalogs: Social Re-Working of Genre on LibraryThing," *Proceedings of the ACM on Human-Computer Interaction* 5, no. 29 (April 13, 2021): 1–29.

| Change in Rank | Shelves That Rose Most in Rank | Change in Rank | Shelves That Fell Most in Rank (or Rose Least) |
| --- | --- | --- | --- |
| +0.75 | read-in-school | +0.06 | to-read |
| +0.73 | school-reading | +0.04 | bloomsbury |
| +0.72 | william-shakespeare | +0.03 | stream-of-consciousness |
| +0.71 | thriller-mystery | +0.01 | virago-modern-classics |
| +0.71 | chick-lit | -0.03 | lit-crit |
| +0.70 | classics-read | -0.03 | literary-criticism |
| +0.69 | the-classics | -0.03 | virago |
| +0.69 | realistic-fiction | -0.05 | d-h-lawrence |
| +0.69 | childrens-lit | -0.05 | criticism |
| +0.68 | great-american-read | -0.08 | joyce |

**Table 8.** Shelves that rose and fell the most in rank from SC to Goodreads.

In sum, our work on lists and shelves shows that even the sustained prominence of authors and books that readers today associate with canonical modernism does not match the status that those authors and books enjoyed among SC members. While literary status can indeed be very durable, our results suggest the early heights of popularity reached by a few authors within the relatively small and distinctive SC group have not been sustained even at the highest end of those evaluated by the larger, later, and more diverse Goodreads community.

**CHRONOLOGICAL COMPARISON**

In the previous section we noted that the most extreme changes in popularity appeared to be correlated with year of publication. Here we provide a more complete analysis of this relationship. We focus on books published between 1800 to 1940, for which we have the most reliable data. We find that books that are more popular in Goodreads are spread roughly uniformly over the time period, while the works more popular in SC are concentrated between 1910 and 1940.

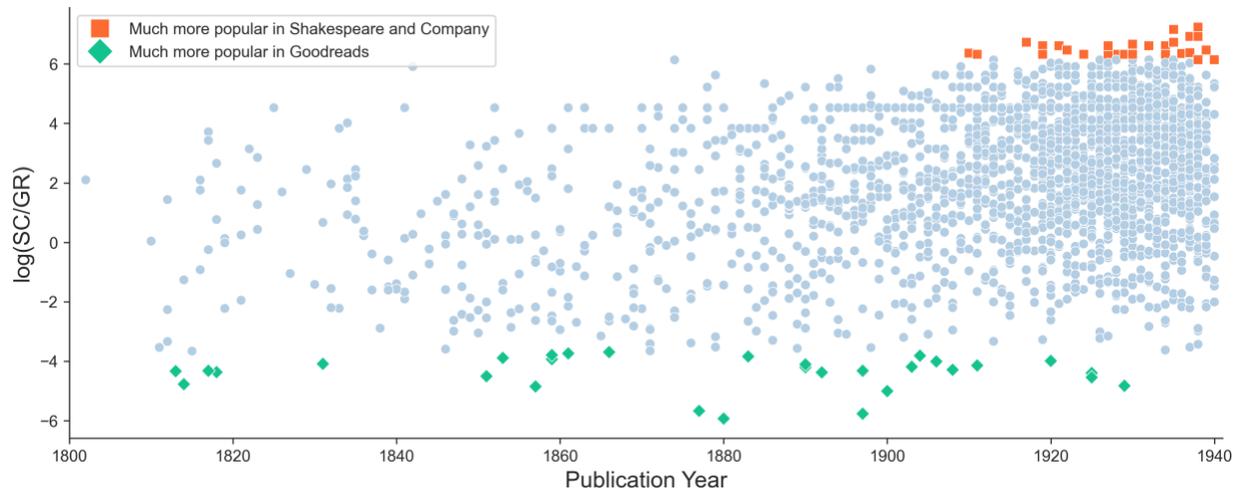

**Fig. 3.** Relative popularity of titles across SC and Goodreads for the 2225 titles published in the nineteenth and twentieth centuries with non-zero popularity in both datasets. The y-value for each work is the proportion of total borrows in SC accounted for by the work, divided by the proportion of total reviews in Goodreads accounted for by the work. Positive y-values mean that a book was more popular in SC. Negative y-values mean a book is more popular in Goodreads. The most relatively popular books in SC are all from the twentieth century, while the most relatively popular books in Goodreads are drawn more uniformly across the nineteenth and early twentieth centuries. (For reference, the top five SC titles by this metric are *The Midas Touch* [1938] by Margaret Kennedy, *Ripeness is All* [1935] by Eric Linklater, *This is Mr. Fortune* [1938] by H. C. Bailey, *The White Horses of Vienna* [1937] by Kay Boyle, and *The Washington Legation Murders* [1935] by F. Van Wyck Mason. The five Goodreads titles are *Little Women* [1880], *Dracula* [1897], *Anna Karenina* [1877], *The Wonderful Wizard of Oz* [1900], and *Madame Bovary* [1857].)

The top hundred works in Goodreads show much greater variation in publication date than do the top hundred works in SC. Of the books that appear in both SC and Goodreads, older titles were less popular in SC but have a greater chance of still being read in Goodreads. Examples include *Wuthering Heights* (1847), *Moby-Dick* (1851), and *Tess of the D'Urbervilles*

(1891). Earlier works that have most fallen in popularity include *The Egoist* (1879), *Esther Waters* (1894), and *The Plays of Bernard Shaw* (1898).

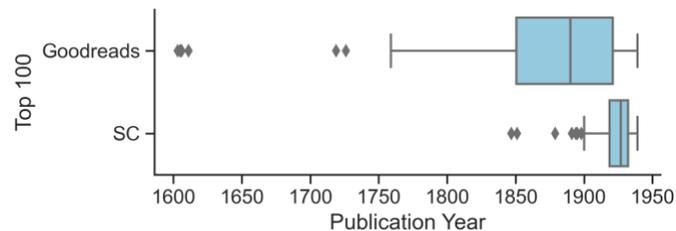

**Fig. 4.** Box plots of publication years for the most popular books in each dataset, truncated to remove lowest outliers for Goodreads.

**COMPARING CONTEMPORARY LITERATURE TO US BESTSELLERS**

Although we observe that the works that have fallen most in popularity tend to be contemporary literature of the 1920s and 1930s, SC readers were nevertheless reading many works that have remained popular. In order to contextualize their reading patterns, we compare to one available source, lists of the ten bestselling novels in the US reported by *Publishers Weekly*.[4] Focusing on works published from 1920–1929, we compare the number of Goodreads ratings for the ninety-seven distinct US bestsellers (three works were bestsellers in two consecutive years) to the hundred most-borrowed SC works from that decade. The lists are measuring different events (purchases by year vs. borrowing over two decades) and the SC values include some works that are not novels and would not be counted, but they provide a point of comparison. The US bestsellers are mostly unfamiliar to modern readers, with a median of eighty-two ratings and a mean of 7,050. Only six works have more than ten thousand ratings, with *The Age of Innocence* and *All Quiet on the Western Front* skewing the mean. In contrast, twenty-six of the top SC works have more than ten thousand ratings, with a median of 1,030 and a mean of 64,151 (skewed by *The Great Gatsby* at #84). Moreover, many of the books borrowed most often by SC patrons remain popular today, with fifteen of the top twenty-five SC works having more than 10,000 ratings by Goodreads users. The lists of bestsellers and of most-borrowed SC titles are not similar, with only eight works in common, of which five are by Sinclair Lewis. In short, popularity among SC readers was a significantly better predictor of enduring literary status than was commercial success at the time of publication.

---

[4] "*[Publishers Weekly list of bestselling novels in the United States in the 1920s](#)*," *Wikipedia*, November 2, 2022.

**COMPARING READING PATTERNS OF POPULAR BOOKS**

Reception is defined not just by the frequency with which a book is read, but also by whom and in what contexts it is read. In the previous section, we measured the popularity of individual books, but we can also consider patterns in reading behavior between books. There is a limit to what we can learn from individual book popularity alone, while book co-reading patterns provide a more detailed and nuanced picture. Additionally, such patterns allow us to study co-reading across the entire community of readers rather than limiting ourselves to which books were checked out by a single reader. To compare co-reading patterns between the two periods, we further restrict our dataset of matched titles to 1,685 books for which we have user information from the UCSD Book Graph.[5] This dataset contains more user information than is easily available from Goodreads. Each dataset induces a network: in SC, two books are connected if they were borrowed by the same member; in Goodreads, books are connected if they were reviewed by the same user. We restrict our network analysis to the 1,511 books that are connected in both datasets.

---

[5] Mengting Wan and Julian McAuley, "Item Recommendation on Monotonic Behavior Chains," in *Proceedings of the 12th ACM Conference on Recommender Systems* (Sept. 2018): 86–94. Mengting Wan

We can compare the neighbors of each book in the two communities. If, for a given book, these neighborhoods of books are similar in the two graphs, we have evidence that the book was read in similar textual company by the members of the two communities. If, in contrast, these neighbor graphs are markedly different, we have evidence that a book has been received in the company of different books. For example, a work might be seen as representing a specific genre in SC, but modern readers might consider the same work to be defined primarily by its prestige or "classic" status.

We operationalize similarity between reading patterns across the two graphs by representing the neighbors of each book as a numeric vector. For each book we define two vectors, the first representing reading patterns in SC and the other representing reading patterns in Goodreads. A given book's vector is indexed by books, where each entry is proportional to the number of readers who interacted with the given pair of books and the vectors are $l_1$-normalized into distributions. We use Jensen-Shannon divergence as a standard method for comparing each book's two vectors.[6]

---

et al., "Fine-Grained Spoiler Detection from Large-Scale Review Corpora," *Proceedings of the 57th Annual Meeting of the Association for Computational Linguistics* (July 2019): 2605–10.

[6] Kent K. Chang and Simon DeDeo, "Divergence and the Complexity of Difference in Text and Culture," *Journal of Cultural Analytics* 5, no. 2 (2020): 1–36.

A persistent challenge in this work is that the number of borrowing events in SC is much smaller than the number of reviews on Goodreads. Since most books are borrowed rarely, many of the interactions between books are even more sparse and potentially noisy. We calculate Jensen-Shannon divergence between each book's distributions over co-occurring books after adding a small constant (0.01) to all vector entries to make the problem well-posed with a uniform prior. We limit this analysis to only the 216 books in the top quartile of popularity in both datasets (that is, books borrowed at least four times in SC and rated at least 2,600 times in the UCSD Book Graph) in order to focus on reading patterns of enduringly popular books.

Books with the most similar distributions across SC and Goodreads are listed in Table 9. These popular books were popular in the same way in SC and Goodreads: often extremely popular and read in conjunction with the same sets of other popular books.

| Rank | Divergence | SC neighbors | GR neighbors | Title | Author |
| --- | --- | --- | --- | --- | --- |
| 1 | 0.3787 | 744 | 390 | *A High Wind in Jamaica* | Richard Hughes |
| 2 | 0.3855 | 662 | 590 | *As I Lay Dying* | William Faulkner |
| 3 | 0.3863 | 704 | 455 | *Light in August* | William Faulkner |
| 4 | 0.3909 | 526 | 407 | *Barchester Towers* | Anthony Trollope |
| 5 | 0.3918 | 915 | 581 | *A Portrait of the Artist as a Young Man* | James Joyce |
| 6 | 0.3928 | 656 | 428 | *A Handful of Dust* | Evelyn Waugh |
| 7 | 0.3928 | 872 | 587 | *A Farewell to Arms* | Ernest Hemingway |
| 8 | 0.3930 | 681 | 357 | *The Way of All Flesh* | Samuel Butler |
| 9 | 0.3935 | 750 | 317 | *The Rainbow* | D. H. Lawrence |
| 10 | 0.3960 | 862 | 349 | *Sanctuary* | William Faulkner |

**Table 9.** The popular books with the lowest Jensen-Shannon divergence have the most similar distributions over co-occurring books.

Even the books with the most similar reading patterns nevertheless have noticeably different neighbors in the two datasets. For example, in both datasets, *Light in August* by William Faulkner is read by people who also read Hemingway and other works by Faulkner. But in SC, top neighbors include now-less-popular works such as *Sanctuary* and *Men Without Women*. In Goodreads, the neighborhood instead includes *The Sound and the Fury* and *The Sun Also Rises*. The contemporary but now-less-read *The Years* by Virginia Woolf appears in SC, while the much older but now-more-popular *Jane Eyre* is more related in Goodreads.

| Rank | Shakespeare and Company | Goodreads |
|---|---|---|
| 1 | *Sanctuary* by William Faulkner | *As I Lay Dying* by William Faulkner |
| 2 | *Men Without Women* by Ernest Hemingway | *The Great Gatsby* by F. Scott Fitzgerald |
| 3 | *As I Lay Dying* by William Faulkner | *The Sound and the Fury* by William Faulkner |
| 4 | *A Farewell to Arms* by Ernest Hemingway | *Jane Eyre* by Charlotte Brontë |
| 5 | *The Years* by Virginia Woolf | *The Sun Also Rises* by Ernest Hemingway |

**Table 10.** The books most frequently interacted with by people who read *Light in August* by William Faulkner, a book with high neighbor similarity across SC and Goodreads.

The books with the highest divergence have the most dissimilar distributions across SC and Goodreads, often due to differences in popularity. Some of these books, such as *The Picture of Dorian Gray*, *Hamlet*, and *The Age of Innocence*, have become "classics" in Goodreads and are read in the company of other "classics." Others were popular in SC but have faded in Goodreads popularity. This category includes both popular books and lesser-known works of still-popular authors, like Oscar Wilde's *Salomé* and Upton Sinclair's *Oil!*

| Rank | Divergence | SC neighbors | GR neighbors | Title | Author |
| --- | --- | --- | --- | --- | --- |
| 1 | 0.5479 | 185 | 210 | *The Varieties of Religious Experience* | William James |
| 2 | 0.5447 | 118 | 146 | *Gitanjali* | Rabindranath Tagore |
| 3 | 0.5413 | 276 | 68 | *The Picture of Dorian Gray* | Oscar Wilde |
| 4 | 0.5383 | 152 | 277 | *Salomé* | Oscar Wilde |
| 5 | 0.5379 | 202 | 212 | *Oil!* | Upton Sinclair |
| 6 | 0.5374 | 120 | 114 | *Selected Poems* | Ezra Pound |
| 7 | 0.5374 | 372 | 556 | *Hamlet* | William Shakespeare |
| 8 | 0.5309 | 41 | 190 | *The Guermantes Way* (*À la recherche du temps perdu* 3) | Marcel Proust |
| 9 | 0.5260 | 78 | 611 | *The Age of Innocence* | Edith Wharton |
| 10 | 0.5230 | 252 | 152 | *Untouchable* | Mulk Raj Anand |

**Table 11.** The popular books with the highest Jensen-Shannon divergence have the most dissimilar distributions over co-occurring books.

Books with high divergence have very different top neighbors, as can be seen in the table below for Wilde's *The Picture of Dorian Gray*, one of the books with the most dissimilar distributions. We see in this case a clear example of a book that, for SC readers, was a living contemporary novel, but today has become primarily a classroom classic.

| Rank | Shakespeare and Company | Goodreads |
| --- | --- | --- |
| 1 | *Tobacco Road* by Erskine Preston Caldwell | *The Great Gatsby* by F. Scott Fitzgerald |
| 2 | *The Way of All Flesh* by Samuel Butler | *Jane Eyre* by Charlotte Brontë |
| 3 | *The Maurizius Case* by Jakob Wassermann | *Brave New World* by Aldous Huxley |
| 4 | *The Idiot* by Fyodor Dostoyevsky | *Emma* by Jane Austen |
| 5 | *The Charterhouse of Parma* by Stendhal | *Crime and Punishment* by Fyodor Dostoyevsky |

**Table 12.** The books most frequently interacted with by people who read Oscar Wilde's *The Picture of Dorian Gray*, a book with low neighbor similarity across SC and Goodreads.

**COMPARING NETWORK ROLES OF POPULAR BOOKS**

In the previous sections, we examined books individually and in the context of their network neighbors. In this final section, we compare the properties of the two readership networks in their entirety. One method for describing a network is to identify its core-periphery structure.[7] We divide the SC network into books that were highly connected to each other through shared readership (the core) and other books that were connected to the core but not to each other (the periphery) using an algorithm that infers a distribution over five nested layers for each node.[8] The innermost layer (layer 1) represents the books assigned to the core while the outermost layer (layer 5) includes books assigned to the periphery. The layers between the core and periphery (layers 2–4) suggest a book's position in a range from core to periphery. We find that books with high average coreness in the SC network tend to have been published in the 1920s and 1930s, but we also find that some of these books were rarely borrowed in SC.

---

[7] Stephen P. Borgatti and Martin G. Everett, "Models of core/periphery structures," *Social Networks* 21, no. 4 (2000): 375–95.

[8] Ryan J. Gallagher, Jean-Gabriel Young, and Brooke Foucault Welles, "A clarified typology of core-periphery s networks," *Science Advances* 7, no. 12 (2021): 1–11.

| Year | Title | Author | Coreness | Borrower Events |
|---|---|---|---|---|
| 1930 | *Diary of a Provincial Lady* | E. M. Delafield | 0.905 | 13 |
| 1937 | *Famine* | Liam O'Flaherty | 0.905 | 6 |
| 1931 | *Back Street* | Fannie Hurst | 0.905 | 6 |
| 1939 | *Beware of Pity* | Stefan Zweig | 0.905 | 5 |
| 1928 | *Debonair: The Story of Persephone* | G. B. Stern | 0.905 | 3 |
| 1926 | *The Sun Also Rises* | Ernest Hemingway | 0.9 | 29 |
| 1915 | *The Rainbow* | D. H. Lawrence | 0.9 | 16 |
| 1925 | *Manhattan Transfer* | John Dos Passos | 0.895 | 26 |
| 1925 | *The Great Gatsby* | F. Scott Fitzgerald | 0.895 | 10 |
| 1885 | *Marius the Epicurean* | Walter Pater | 0.895 | 7 |

**Table 13.** The books with highest coreness in the SC network.

The most core books range in popularity within SC, but were consistently read by SC's top two readers, Alice M. Killen and France Emma Raphaël. Killen and Raphaël were the most prolific borrowers, with 1,480 and 999 borrow events respectively (compared to a median of just nine borrow events across other SC members). Both women were SC members for decades (Killen from 1922–1940, Raphaël from 1920–1942). Killen borrowed 377 and Raphaël borrowed 283 of the books shared between SC and Goodreads; 134 of these books overlap between the two readers.

Though relatively little information about either woman survives today, we know that Killen published a graduate thesis on gothic literature in 1915 and kept in touch with Sylvia

Beach during Killen's vacations to Ireland.[9] Raphaël was an artist and sculptor, and amanuensis for James Joyce.[10] Between 1933–37, Raphaël transcribed Joyce's notebooks; her transcriptions aided Joyce in the drafting of *Finnegans Wake*. Some scholarship on Joyce discusses Raphaël's role in the composition of *Finnegans Wake*, given the prevalence of mistakes Raphaël apparently made during transcription (e.g., "colour of hair & voice" transcribed as "Colonel if have a voice").[11] However, Raphaël's errors "may have delighted Joyce more than [they] disturbed him" (Tigges, "The Fallacy of First-Draft Authority," 835). Because Killen and Raphaël individually had high book-borrowing rates and read many of the same books, together their readership creates the dense core of books identified by the core-periphery model.

---

[9] Alice M. Killen, *Le roman terrifiant ou, roman noir de Walpole à Anne Radcliffe et son influence sur la littérature française jusqu'en 1840* (Paris: Champion, 1924)   Alice M. Killlen, "Sylvia Beach Papers," n.d., C0108, box 23, folder 26, Manuscripts Division, Department of Special Collections, Princeton University Library, Princeton, NJ.

[10] "France Emma Raphaël," ArtNet, December 28, 2020, artnet.com/artists/france-emma-raphaël/.

[11] Wim Tigges, "James Joyce, Mme Raphael, Danis Rose, and the Fallacy of First-Draft Authority," *James Joyce Quarterly* 35, no. 4 (1998): 834–40; Danis Rose, "The Raphael transcriptions: 'flexionals' into 'fleas and snails,'" in *The Textual Diaries of James Joyce*, ed. Danis Rose (Dublin: The Lilliput Press, 1995), 161–89.

| Core-Periphery Layer | Total Books | Killen | Raphaël |
|---|---|---|---|
| 1 | 231 | 231 | 124 |
| 2 | 146 | 146 | 10 |
| 3 | 197 | 0 | 149 |
| 4 | 371 | 0 | 0 |
| 5 | 566 | 0 | 0 |

**Table 14.** Number of SC books assigned to each layer in comparison to the number of books in that layer read by Killen or Raphaël. These numbers only include books that were co-reviewed or co-borrowed in both Goodreads and SC.

Regardless of whether a book was unpopular among SC and Goodreads readers (*Debonair: The Story of Persephone*), gained popularity over time (*The Great Gatsby*), or was popular both in SC and Goodreads (*The Sun Also Rises*), the most "core" books were borrowed by both Killen and Raphaël. Indeed, other network centrality measures, including betweenness centrality, degree centrality, and eigenvector centrality, also rank books borrowed by both Raphaël and Killen highest, though high-centrality books tend to have been more popular with SC readers in general.

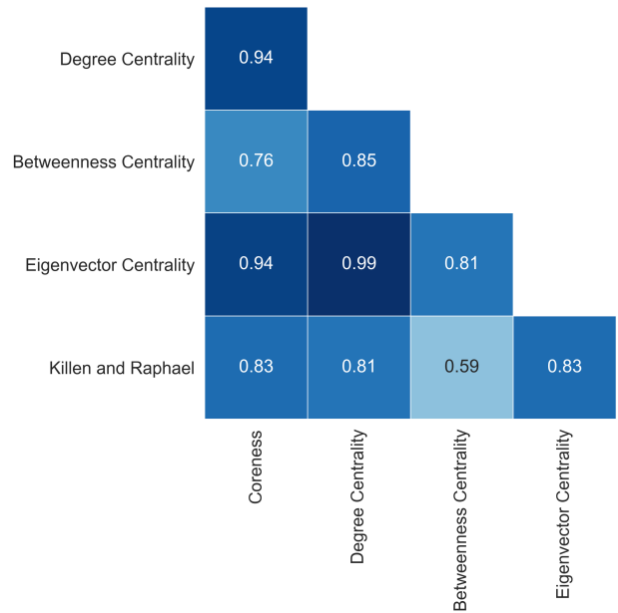

**Fig. 5.** Spearman correlation coefficients between network coreness and centrality measures and books read by Killen and Raphaël. "Killen and Raphaël" refers to books checked out by Killen and/or Raphaël.

Given their twenty-year overlap at SC, adjacent addresses, and high overlap of books borrowed, we can speculate that the two women likely shared and discussed SC books (fig. 6). Killen and Raphaël consecutively borrowed otherwise unpopular books (Stefan Zweig's *Beware of Pity*; G. B. Stern's *Debonair: The Story of Persephone*) within five days of each other. From 1934–1940, a third reader, Françoise de Marcilly, was also likely associated with Killen and Raphaël, though Marcilly's reading habits did not have the same impact on the SC network. For example, on April 8, 1936, Marcilly returned Charles Morgan's *Sparkenbroke*. On the same day, Raphaël borrowed *Sparkenbroke*, returning it on April 15; Killen borrowed the book five days later. The three women also co-read other books that were unpopular in SC, such as H. M. Tomlinson's *The Sea and the Jungle*, which was borrowed only five times; Killen, Raphaël, and Marcilly account for three of these events (the other two borrowers were Ernest Hemingway and Giorgio Joyce, James Joyce's son).

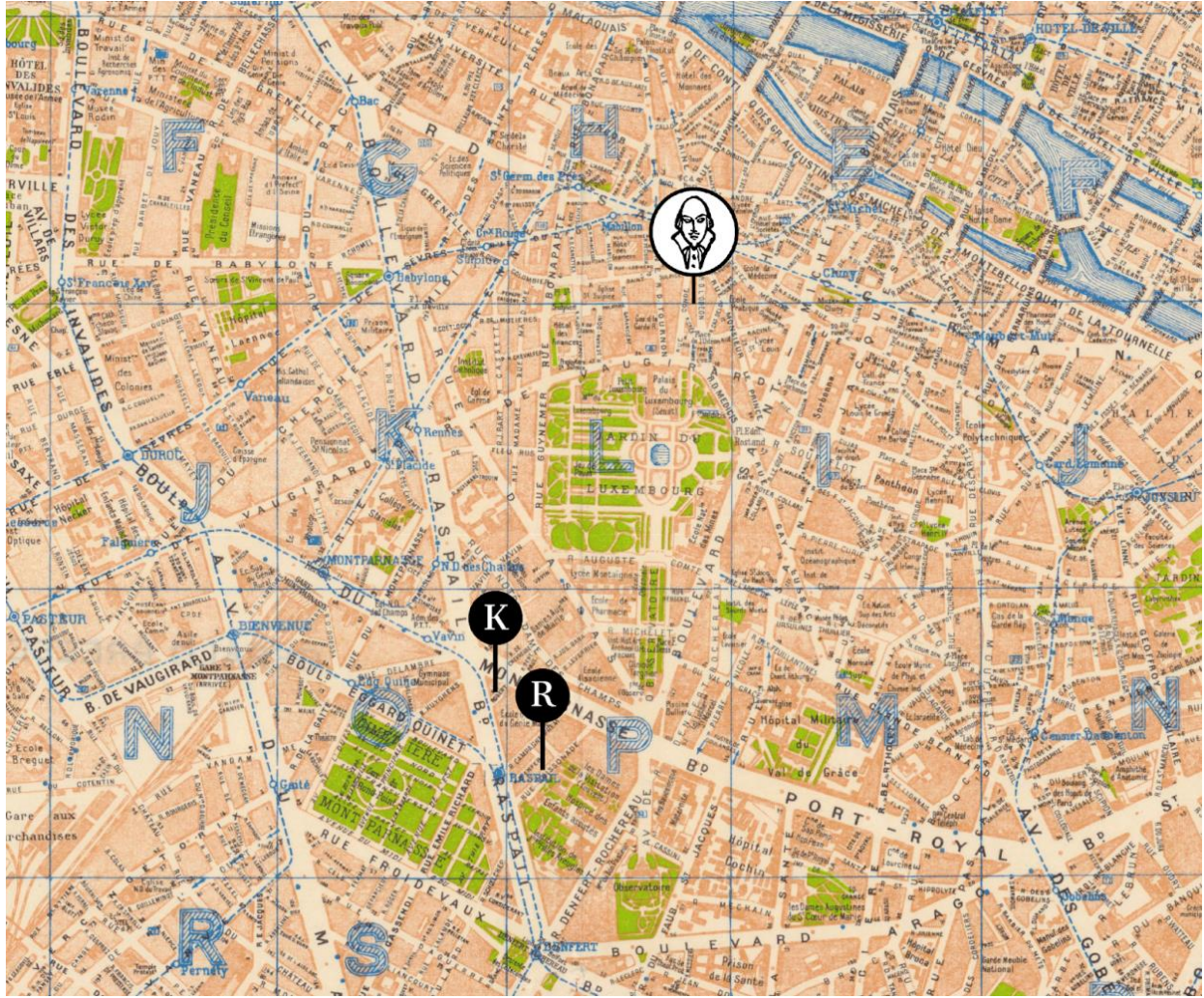

**Fig. 6.** Addresses listed in Paris for Killen (8 rue Léopold Robert, Paris 14e), Raphaël (40 rue Boissonade, Paris 14e), and SC (12 rue de l'Odéon, Paris 6e). Raphaël's extant membership cards also locate her on the same block, at 18 rue Boissonade, from 1929–1935, and another address, further away from SC, in 1926. Raphaël and Killen have addresses listed outside of Paris, in Southern France and Ireland, respectively.

Of all books borrowed at SC, Killen and Raphaël read books published in a narrower range of years, usually preferring contemporary works. The median publication date of all books in SC is 1921, while the median publication year of books borrowed by Killen was 1926 and by Raphaël, 1928. Comparing the popularity of their borrowed books across SC and Goodreads, we find that the books Killen and Raphaël borrowed usually fell in popularity, though this result is influenced by their preference for contemporary works (which, as previously observed, tend to have fallen in popularity more than have earlier titles). The members whose borrowed books rose most in popularity (Isabelle Zimmer, Dorothy Dudley Harvey, and Nathalie Sarraute) read comparatively few books relative to Killen and Raphaël, who read widely across contemporary works available at SC.[12] Characteristics of Killen and Raphaël's borrowing habits distinguish the pair from the average SC reader, given the scale of their borrowing, their tendency to borrow contemporary work, and the lessened popularity of books they borrowed over time. Still, the borrowing attributes that distinguish them also emphasize their similar reading habits, strengthening the likelihood that they influenced each other's reading by sharing and recommending books.

When applied to the SC network, our network significance measures behave surprisingly, revealing more about two prominent readers than about the network as a whole. While network analysis measures such as centrality and core-periphery structure usually allow for a birds-eye-view of a network, in this case they magnify the readership of two prolific book-borrowers. Though there is little extant information about either Killen or Raphaël, their prominent position in the SC book network emphasizes the frequency and longevity of their book-borrowing from SC. They, and their relationship, merit further study.

---

[12] We identify SC members whose borrowed books rose most in rank by first filtering to only those members who had at least ten borrowing events and then, for each book they borrowed, calculating that book's change in popularity rank from SC to Goodreads. We then take the mean of these scores for each SC member.

**CONCLUSION**

By putting an interwar Parisian lending library in conversation with a large, contemporary social reading website, we can compare reception over time, place, and audience. Books that were more popular in SC tend to be experimental and written at the time SC was active, while those that became more popular on Goodreads are older and now marked by the crowd as "classics" and "school" books. By looking at patterns of co-readership, we find variability in the process of canonization. Some works, such as those by Faulkner and Hemingway, largely retain their original context—they are read today in much the same literary company they enjoyed in SC's Paris. Others are more disconnected from their contemporary literature, becoming "timeless" as readers associate them with important (or widely taught), temporally varied peers.

The approaches we have employed provide a data-driven window into both the prescience and influential taste of a historical community, as well as its passing obsessions. These results offer insights, but also highlight the limitations of the datasets and the difficulties of making meaningful comparisons between them. While there remain questions about the role of computational approaches in the study of literature itself, we believe the use of digital methods for the study of the reception of literature is both necessary and powerful. The Goodreads dataset is fundamentally mediated by computation; it would be impossible to approach in any other way. Even at the smaller scale of SC, we have enough data (thousands of records about thousands of books; an induced network of millions of possible interactions) from which computational analysis can reveal unexpected details. But in stretching the limits of quantitative comparison, we nevertheless reach a point where our methods tell us less about literature and reception, and more about two neighbors walking through the park to pick up the week's new novel.

**Maria Antoniak**

mariaa@allenai.org

Maria Antoniak is a Young Investigator at the Allen Institute for Artificial Intelligence on the Semantic Scholar team. Her research is in natural language processing and cultural analytics, and she has a PhD from Cornell University in Information Science and a master's degree from the University of Washington in Computational Linguistics. She has worked with Microsoft Research, Twitter, Facebook, and Pacific Northwest National Laboratory.

**David Mimno**

mimno@cornell.edu

David Mimno is an Associate Professor in the department of Information Science at Cornell University. He holds a PhD from UMass Amherst and was previously the head programmer at the Perseus Project at Tufts and a researcher at Princeton University. His work has been supported by the Sloan foundation, NEH, and NSF.

**Rosamond Thalken**

ret85@cornell.edu

Rosamond Thalken is a PhD student in Information Science at Cornell University. After earning her MA in Literature at Washington State University, she was an Instructor in Digital Technology and Culture and Women's, Gender, and Sexuality Studies at WSU. She co-authored the second edition of *Text Analysis with R: For Students of Literature*.

**Melanie Walsh**

melwalsh@uw.edu

Melanie Walsh is Assistant Professor in the Information School at the University of Washington. Her research is in the areas of data science, digital humanities, cultural analytics, and literature. She is the author of a free online programming textbook, *Introduction to Cultural Analytics & Python*. She is also co-editor of the Post45 Data Collective, a peer-reviewed, open-access repository for literary and cultural data, and she co-directs the BERT for Humanists project.

**Matthew Wilkens**

wilkens@cornell.edu

Matthew Wilkens is Associate Professor of Information Science at Cornell, where he works on quantitative methods for literary and cultural history. His current research also explores computational analysis of online health communities. He serves on the editorial board of the *Journal of Cultural Analytics*, directs the Textual Geographies project, and co-directs the AI for Humanists project. He is the author of *Revolution: The Event in Postwar Fiction*.

**Gregory Yauney**

gyauney@cs.cornell.edu

Gregory Yauney is a PhD candidate in computer science at Cornell University. He is interested in machine learning theory and digital humanities.